\documentclass[10pt,twocolumn,letterpaper]{article}

\usepackage{cvpr}
\usepackage{times}
\usepackage{epsfig}
\usepackage{graphicx}
\usepackage{amsmath}
\usepackage{amssymb}
\usepackage{bbm}

\usepackage{booktabs}
\usepackage[table]{xcolor}
\usepackage{multirow}
\usepackage{cite}

\usepackage[pagebackref=true,breaklinks=true,letterpaper=true,colorlinks,bookmarks=false]{hyperref}

\cvprfinalcopy 

\def\cvprPaperID{523} 

\def\eg{\emph{e.g}\onedot} 
\def\ie{\emph{i.e}\onedot}

\def\etc{\emph{etc}\onedot} 
\def\etal{\emph{et al}\onedot}
\def\etal{$et\ al.$}
\newcommand{\argmin}{\operatornamewithlimits{argmin}}

\usepackage{cleveref}[2012/02/15]
\crefformat{footnote}{#2\footnotemark[#1]#3}

\def\y{{{\mathbf{y}}}}
\def\t{\mathcal{T}}
\ifcvprfinal\fi
\begin{document}

\title{Multi-Instance Visual-Semantic Embedding}

\author{Zhou Ren$^1$\ \ \ \ \ Hailin Jin$^2$\ \ \ \ \ Zhe Lin$^2$\ \ \ \ \ Chen Fang$^2$\ \ \ \ \ Alan Yuille$^1$\\
$^1$University of California, Los Angeles\ \ \ \ \ \ \ $^2$Adobe Research
}

\maketitle

\begin{abstract}
Visual-semantic embedding models have been recently proposed and shown to be effective for image classification and zero-shot learning, by mapping images into a continuous semantic label space. Although several approaches have been proposed for single-label embedding tasks, handling images with multiple labels (which is a more general setting) still remains an open problem, mainly due to the complex underlying corresponding relationship between image and its labels. In this work, we present Multi-Instance visual-semantic Embedding model (MIE) for embedding images associated with either single or multiple labels. Our model discovers and maps semantically-meaningful image subregions to their corresponding labels. And we demonstrate the superiority of our method over the state-of-the-art on two tasks, including multi-label image annotation and zero-shot learning. 

\end{abstract}
\vspace{-5pt}

\section{Introduction}
Image classification is a fundamental problem in computer vision. The classic approach tries to categorize images into a fixed set of classes by training a multi-class classifier. And continuous efforts have been made in building image classification systems with larger-scale image datasets as well as a broader coverage of visual classes. However, due to complex relationships among the concepts (\eg, hierarchical, disjoint, etc.), it is hard to define a perfect classifier encoding all those semantic relations \cite{JiaDengECCV14}. Also, since the set of classes is predefined, such systems need to be re-trained whenever a new visual entity emerges.

To address these shortcomings, visual-semantic embedding methods \cite{DeviSE, NoriuziICLR14} are introduced which leverage semantic information from unannotated text data to learn semantic relationships between labels, and explicitly map images into a rich semantic embedding space (in such space, certain semantic relationships are encoded, \eg, \emph{sun} and \emph{sunrise} may locate at close positions. And images from these two classes may share some common visual appearance.). By resorting to classification in the embedding space with respect to a set of label embedding vectors, visual-semantic models have shown comparable performance to the state-of-the-art visual object classifiers and demonstrated \emph{zero-shot learning} capability, \ie, correctly predicting object category labels for unseen categories (which validates the ability to address the shortcomings mentioned before).

\begin{figure}
  \centering
  \includegraphics[width=0.48\textwidth] {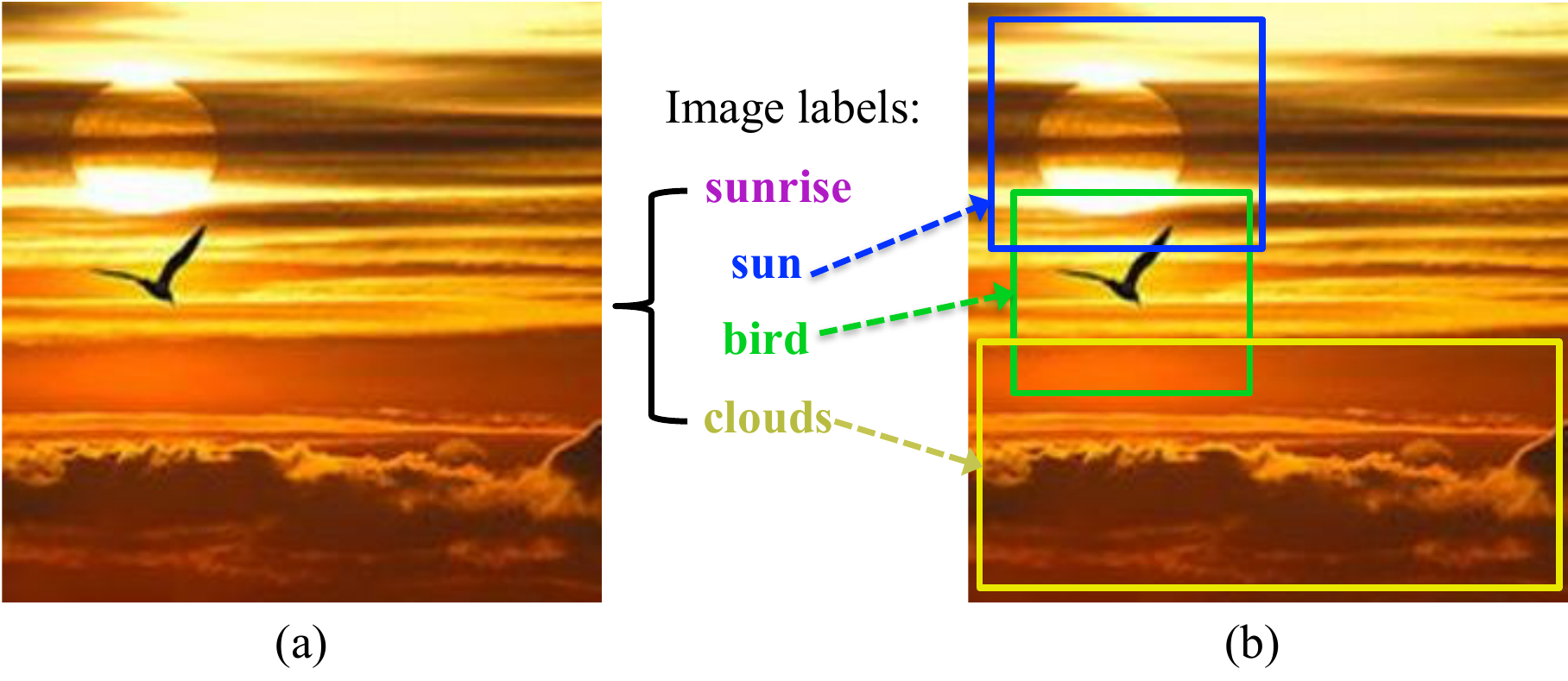}
\caption{ (a) An example of an image with multiple labels, which are listed in the middle. (b) We observe that different labels may correspond to various image subregions, but not necessarily the whole image, such as the labels \emph{clouds}, \emph{sun}, \emph{bird} are associated with the subregions in the bounding boxes.}
\label{FigIntro}
\vspace{-5pt}
\end{figure}

Although visual-semantic embedding models have shown impressive preliminary results on visual tasks for images with single labels, no attempts have been made on optimizing it for multi-label classification (or annotation) problem. And it is of great importance to develop such a model due to the following reasons. Firstly, real-world images are often associated with multiple description labels. Existing visual-semantic embedding models are developed for single-label images only. Secondly, it is nontrivial to extend a single-label visual-semantic embedding model to a multi-label one. The implicit assumption that all image labels correspond to the whole image does not hold for multi-label cases. For a typical multi-label image, some labels may correspond to image subregions, instead of the whole image. For example, as shown in Fig.\ref{FigIntro}, only the scene label \emph{sunrise} corresponds to the whole image, and all other label concepts correspond to specific subregions in the image as shown in the bounding boxes.

To address these problems, we present Multi-Instance visual-semantic Embedding model (MIE) to embed images with single or multiple labels. To this end, we learn an embedding model that discovers and maps semantically-meaningful image subregions to their corresponding labels in the semantic space. To automatically generate semantically-meaningful regions for each label, we utilize region proposals method \cite{GOPECCV14} to construct the image subregion set. Then, we infer the best-matching region for each label jointly to build region-to-label correspondence. This model allows us to avoid embedding an image to distant labels, and generate a image subregion ``instance" for each label concept as best as it can.

The effectiveness of our model is evaluated on two tasks. In the multi-label image annotation task, our model can correctly annotate images with multiple labels (which outperforms state-of-the-art method by 4.5\%) and discover the semantically-meaningful subregion associated with each label (which is a distinct advantage of our method). And for the zero-shot learning task, the superiority of our model is further validated by comparing with the baseline embedding model on new category classification. 
\vspace{-1pt}

\section{Related Work}
We briefly outline connections and differences to three related lines of research.
\vspace{5pt}

\noindent \textbf{Multi-modal Embedding models}
\vspace{2pt}

Multi-modal embedding models relate information from multiple sources, such as images and texts.
Attributes are a common way to encode the presence and absence of a set of visual properties within an object instance. In attribute-based approaches \cite{attribute1, relativeattr11}, the semantic space was constructed by representing each class label by a vector of attributes. WSABIE \cite{wsabie11} proposed a loss function specifically designed for image annotation task to embed images with the associated multiple labels together. While these embedding approaches effectively capture richer label information than multi-class classifiers, they have not demonstrated the effectiveness of scaling to very large-scale cases. 

To address this problem, visual-semantic embedding models \cite{DeviSE, NoriuziICLR14} were proposed. By leveraging the plenty textual data available on the Internet, scalable and lexically distributed representations of words were learned to capture the semantic meaning among them, \eg, word2vec model \cite{Word2VecNIPS13} and Glove model \cite{glove14}. Thus, instead of manually designing the semantic label space, Frome \etal \cite{DeviSE} and Norouzi \etal \cite{NoriuziICLR14} leveraged such semantic information to learn visual-semantic embedding where semantic relationship between labels was preserved. 

Although those visual-semantic embedding models have shown good scalability on the ImageNet-size dataset with single label, they are not applied to images with multiple labels which is typical for automatic image annotation. 
\vspace{5pt}

\noindent \textbf{Multi-label image annotation}
\vspace{1pt}

Early work for multi-label image annotation focused on learning statistical models based on hand-crafted features, including sparse coding \cite{JianchaoCVPR09}, fisher vectors \cite{FisherVectorCVPR07}, bag-of-words \cite{ChenBOWCVPR12}, etc. For instance, Makadia \etal \cite{MakadiaECCV08} and Guillaumin \etal \cite{Tagprop09} proposed nonparametric nearest-neighbor methods to transfer image labels. WSABIE \cite{wsabie11} demonstrated superior results with a WARP loss based on bag-of-words feature.

Recently, as the learned image representation using deep convolutional neural network (CNN) has shown superior accuracy in various vision tasks \cite{LecunNIPS90, Alexnet12, visualCNN14, googlenet15, vggnet15, fastrcnn15}, Gong \etal \cite{YunchaoICLR14} applied the CNN architecture to the multi-label image annotation problem and achieved the state-of-the-art performance. 
\vspace{4pt}

\noindent \textbf{Zero-shot learning}
\vspace{2pt}
 
Zero-shot learning is commonly used to evaluate the scalability of a system w.r.t. the number of categories, \ie, annotating images with new labels corresponding to previously unseen object categories. 

Early work \cite{marknips09, marcuscvpr11} attempted to solve this problem relying on additional curated semantic information, \eg, \cite{marknips09} used a knowledge base containing descriptions for each class and \cite{marcuscvpr11} used the WordNet hierarchy. Recently, various methods were proposed to learn semantic representation directly from unannotated text data, \eg, visual-semantic embedding models \cite{DeviSE, NoriuziICLR14} and Socher \etal\cite{sochernips13}.

\section{Preliminary on Visual-Semantic Embedding}
We first review the preliminary on visual-semantic embedding and introduce our baseline model.

Given a multi-label image dataset $\mathcal{D}\equiv\{(x_i,\y_i)\}_{i=1}^N$, where each image is represented by a $d$-dimensional feature vector (\eg, a CNN layer output), \ie, $x_i\in \mathbb{R}^d$, and a set of class labels, $\y_i=(y_i^1,...,y_i^t)$, where the label set size $t$ ($t\geq1$) can be varied across images. For generality we denote $\mathcal{X}\stackrel{def}{=}\mathbb{R}^d$. Previous methods formulate the multi-label image classification or annotation problem as multi-class classification, which predefines a fixed set of class labels, \ie, $y_i^l\in \mathcal{Y}\equiv\{1,...,m\}$, and learns an $m$-way classifier or $m$ one-against-all classifiers that classifies images into training labels $(\mathcal{X}\to\mathcal{Y})$. 

\begin{figure*}
  \centering
  \includegraphics[width=0.9\textwidth] {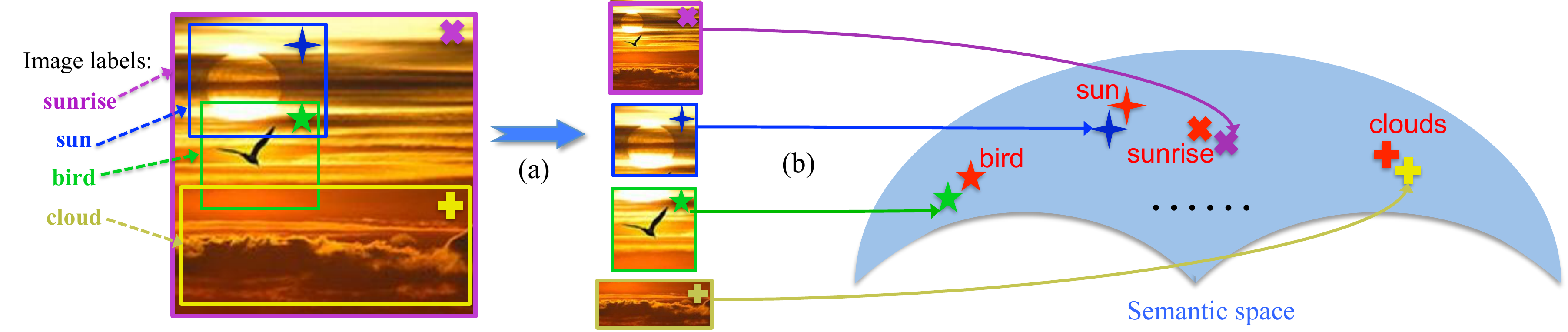}
\caption{ Illustration of our Multi-Instance visual-semantic Embedding model, which is composed of two steps: (a) extract image subregion features; (b) embed each semantically-meaningful image subregion into nearby position of its corresponding label in the semantic space (the red symbols illustrate the embedding of text labels, and symbols of other colors indicate that of different image subregions.). Note: the bounding boxes are for visualization only, and they are not provided in training, better viewed in color. }
\label{Fig2}
\vspace{-10pt}
\end{figure*}

However, this standard classification models lack scalability w.r.t. the number of class labels, and needs to be retrained when any new label emerges. Recently, visual-semantic embedding models \cite{DeviSE} were proposed to overcome this issue for single-label image classification, \ie, instead of defining image labels as a discrete set $\mathcal{Y}$, it aims to learn a continuous semantic space $\mathcal{S}$ which captures the semantic relationship among labels and explicitly learn the mapping function from images to such a space $(\mathcal{X}\to\mathcal{S})$. A training dataset $\{(x_i,s(\y_i))\}$ is constructed to learn an embedding function $f: \mathcal{X}\to\mathcal{S}$ that aims to map $x_i$ to $s(\y_i)$ where $s: \mathcal{Y} \to \mathcal{S}$.

Thus, there are two main problems to explore for a visual-semantic embedding model: 1) how to construct the continuous semantic space $\mathcal{S}$ of image labels; and 2) how to learn the embedding function $f$.

\subsection{Constructing the semantic label space}
\label{glovesec}
The skip-gram text modeling architecture \cite{Word2VecNIPS13, glove14} trained with unannotated text data from the Internet has been demonstrated to provide semantically-meaningful embedding features for text terms. This method is able to learn similar embedding points for semantically related words because of the fact that synonyms are more likely to appear in similar semantic contexts. 

Thus, we utilize the Glove model \cite{glove14} to construct a 300-d text label space $\mathcal{S}$ which embodies the semantic relationship among text labels.

\subsection{Our ranking loss baseline model}
\label{LearnEmbeddingF}
The embedding function for an image-label pair, $f: \mathcal{X}\to\mathcal{S}$, is typically learned with a $L_2$ or a ranking loss function. These loss functions are designed such that, for a given image, its projected point is encouraged to be closer to its corresponding label in the embedding space.
Motivated by the success of the ranking loss in visual-semantic embedding \cite{DeviSE} and multi-label annotation \cite{YunchaoICLR14}, we construct our baseline with the following pairwise hinge-ranking loss: 
\vspace{-17pt}

\begin{small}
\begin{equation}
l_{\text{rank}}(x_i,\y_i)=\!\sum_{j\in \t_+}\sum_{k\in \t_-} \max\left(0, m\!+\!D_{f(x_i),y_j}\!-\!D_{f(x_i),y_k}\right)\!, 
\label{Rankingloss}
\end{equation}
\end{small}
$\!\!$\noindent where $f(\cdot)$ is the embedding function to be learned, $m$ is the margin. For simplicity of notations, we use $y$ instead of $s(y)$ to denote the label vector in $\mathcal{S}$. $\t_+$ and $\t_-$ denote the set of positive labels and negative labels, and $\{y_j\}_{j\in \t_+}$ and $\{y_k\}_{k\in \t_-}$ denote the positive and negative label vectors, respectively. $D_{f(x_i),y_j}$\footnote{\label{note1}In this paper we use squared Euclidean distance to measure distance in the embedding space , \ie, $D_{f(x_i), y_j}=||f(x_i)-y_j||^2_2$.} indicates the distance between the image embedding vector $f(x_i)$ and a label vector $y_j$. 
 
\section{Multi-Instance Visual-Semantic Embedding}
\vspace{-3pt}
Our baseline model presented in Eq.\ref{Rankingloss} is seemingly plausible for multi-label image embedding where each training image is associated with multiple labels. However, there is a problem. Each image $x_i$ may correspond to multiple labels in $\y_i$, but one or more pairs of those labels could be located far away from others in the embedding space $\mathcal{S}$, as shown in Fig.\ref{Fig2}. Trying to push the embedding of a whole image, $f(x_i)$, to be close to multiple distant locations will confuse the embedding function; in the worst case, the image could be mapped to near average position of those label vectors, which might correspond to a totally different concept.

The key observation for overcoming this problem is that different image labels often correspond to different regions in the image. For example, in Fig.\ref{Fig2}, the image has four labels \emph{sunrise}, \emph{clouds}, \emph{sun}, and \emph{bird}. Among them only \emph{sunrise} corresponds to the whole image and other labels correspond to image subregions shown in the bounding boxes. This motivates us to derive a new idea for multi-label embedding in which one can generate multiple different crops (or region proposals) from the image and use the resulting subregion set to match the set of labels in the embedding space. This requires a region-to-label correspondence to be constructed on-the-fly during the learning process.
\vspace{-1pt}

\subsection{Modeling region-to-label correspondence}
\vspace{-1pt}

Based on the above motivation, we propose Multi-Instance visual-semantic Embedding (MIE) to discover the semantically-meaningful image subregions for each label and map them to their corresponding labels in semantic space respectively, as shown in Fig.\ref{Fig2}. The basic loss function to learn MIE embedding is defined as follows (we will improve it in Eq.\ref{MIEloss2}):
\vspace{-8pt}

\begin{small}
\begin{equation}
{l_{\text{MIE}}}(x_i,\y_i)=\!\!\!\sum_{j\in \t_+}\sum_{k\in \t_-} \max(0, m+\min_{c\in \mathcal{C}}D_{f(x_i^c),y_j}\!-\min_{c\in \mathcal{C}}D_{f(x_i^c),y_k}), 
\vspace{-1pt}
\label{MIEloss}
\end{equation}
\end{small}
$\!\!$\noindent where $x_i^c$ indicates a subregion of image $x_i$, $\mathcal{C}$ is the set of all image subregions (we will discuss how to obtain $\mathcal{C}$ in Section \ref{learnMIE}), $\{y_j\}_{j\in \t_+}$ and $\{y_k\}_{k\in \t_-}$ denote the positive and negative label vectors, and $D_{f(x_i^c), y_j}$\cref{note1} indicates the distance between $f(x_i^c)$ and $y_j$.  

As shown in the equation, we simply model the region-to-label correspondence by a $\min$ operation on the distances from a label to all the subregions. This follows our intuition that: 1) each ground truth label should be encouraged to have at least one matching subregion to explain the concept; 2) the subregion with the closest distance to a label is more likely to represent the label.  

\subsection{Optimizing the ranking}
\label{rankoptimize}
However, as stated in \cite{wsabie11}, one limitation of the loss function in Eq.\ref{MIEloss} is that it does not directly optimize the ranking of the predicted labels (instead it optimizes the area under the ROC curve). In order to optimize the ranking of labels, we need to encourage positive (ground truth) labels to have smaller min-distance than most negative labels, \ie, to rank the positive labels on the top of the prediction list. Thus, following \cite{wsabie11}, we give larger penalties to false predictions of ranking positive labels at the bottom, as follows: 
\vspace{-18pt}

\begin{small}
\begin{equation}
\begin{split}
&\widehat{l_{\text{MIE}}}(x_i,\y_i)=\\ &\!\!\!\!\sum_{j\in \t_+}\sum_{k\in \t_-}\!\!w(r_j)\cdot \max(0, m+\min_{c\in \mathcal{C}}D_{f(x_i^c),y_j}\!-\min_{c\in \mathcal{C}}D_{f(x_i^c),y_k}), 
\label{MIEloss2}
\end{split}
\end{equation}
\end{small}
$\!\!$\noindent where $w(\cdot)$ is a weight function defined below, and $r_j$ is the rank of a ground truth label $y_j$ in the prediction:
\vspace{-10pt}

\begin{small}
\begin{equation}
r_j = \sum_{t\neq j} \mathbbm{1}\left(\min_{c\in \mathcal{C}}D_{f(x_i^c),y_t} \leq \min_{c\in \mathcal{C}}D_{f(x_i^c),y_j}\right), 
\label{rankdefine}
\end{equation}
\end{small}
$\!\!\!$\noindent where $\mathbbm{1}(\cdot)$ is the indicator function. As we see, given an image $x_i$, we rank a predicted label $y_t$ according to its distance to all image subregions, \ie, $\min_{c\in \mathcal{C}}D_{f(x_i^c),y_t}$. And we define the weight function $w(\cdot)$ in Eq.\ref{MIEloss2} as follows:
\vspace{-3pt}

\begin{small}
\begin{equation}
w(r) = \begin{cases}
1&\ \ \ \  \mbox{if\ $r< \#(\t_+)$,} \\ 
r&\ \ \ \  \mbox{otherwise}.
\end{cases}  
\label{weightdefine}
\end{equation}
\end{small}

Given an image with $\#(\t_+)$ ground truth labels, if a ground truth label is ranked within top-$\#(\t_+)$ in the prediction list, we give small penalty weights to the loss. While if a ground truth label is not ranked top, \ie, $r_j\geq\#(\t_+)$, we assign much larger weight to it. The intuition for the weighting scheme of ranking optimization in Eq.\ref{MIEloss2} is that it pushes the positive labels to the top, thus pushing meaningful image subregions map closer to their corresponding ground truth labels in the embedding space.

\subsection{Learning multi-instance embedding}
\label{learnMIE}
In the following, we introduce our overall MIE training architecture and explain how we construct $\mathcal{C}$.
\vspace{7pt}

\noindent \textbf{MIE network architecture}
\vspace{3pt}

\begin{figure}
  \centering
  \includegraphics[width=0.5\textwidth] {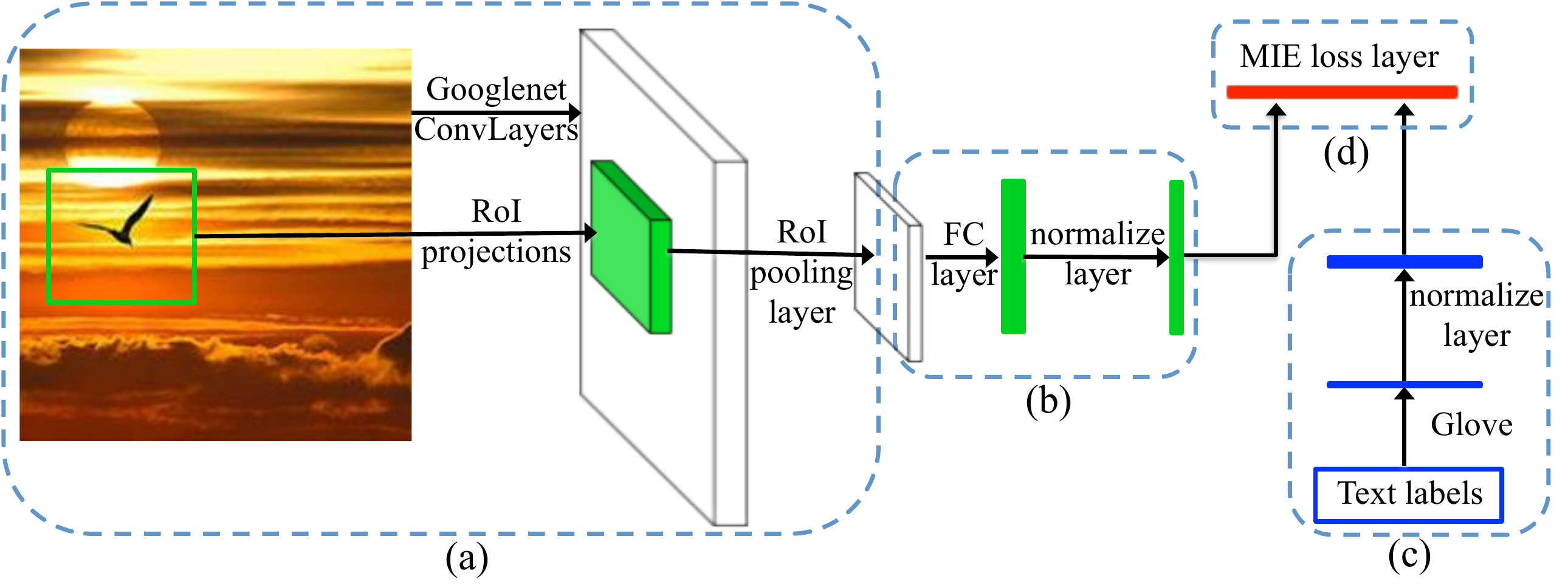}
\caption{ The deep network architecture of MIE model composed of 4 components: (a) subregion (RoI) image features extraction; (b) image features embedding; (c) text features embedding; (d) joint embedding learning guided by the MIE loss layer.}
\label{Fig3}
\end{figure}

Fig.\ref{Fig3} illustrates the overall network architecture of the MIE model. 
As discussed in Section \ref{glovesec}, we utilize the Glove model \cite{glove14} to obtain 300-d text features $\y_i$. And we adopt fully convolutional layers of pretained Googlenet \cite{googlenet15} (including convolution, pooling, and inception layers) to extract 1024-d image features $x_i$\footnote{In principal, MIE can be applied to any image, text representations.}. To learn the embedding from image to semantic space, \ie, $f: \mathcal{X}\to\mathcal{S}$, a fully connected layer is used following the image feature output. Then, $L_2$ normalization layers are added on both image and text embedding vectors, which makes the embeddings of image and text modalities comparable (we tested MIE without the normalization layers and the results were worse). Finally, we add the MIE loss layer on the top to guide the training.

In our MIE model, we need to extract image features for many image subregions within the image, $x_i^c$. For efficiency, we follow the Fast RCNN \cite{fastrcnn15} scheme to extract image subregion features. Namely, given an image $x_i$ and the regions of interests (RoI) $\mathcal{C}$, we pass the image through the fully convolutional network only once, and each subregion $c\in \mathcal{C}$ is pooled into a fixed-size feature map to obtain the final feature vector $x_i^c$. 
\vspace{7pt}

\noindent \textbf{MIE subregion set construction}
\vspace{3pt}

Note that we do not have bounding box annotations for training, thus an essential problem of our MIE model is how to construct the image subregion set $\mathcal{C}$. Inspired by the recent object-proposal-based methods \cite{RCNN14, fastrcnn15} in object detection, we construct the set using a state of the art region proposals method \cite{GOPECCV14} followed by post-processing. 

Since semantically-meaningful subregions of an image do not necessarily contain objects, general object proposal methods that focus on foreground are not suitable here, especially the ones depend on edge information. Hence we use geodesic object proposals \cite{GOPECCV14} since they cover both foreground and background regions reasonably well. Finally, we post-process the subregions to discard regions of too small size or extreme aspect ratio (in our experiments we constraint the subregions' side length to be at least 0.3 of the image and the extreme aspect ratio to be 1:4 or 4:1).



\subsection{Inference with multi-instance embedding}
\label{inference}
Given the trained MIE model, now we introduce how to do inference on a new test image $x^\prime$. 

Firstly, the subregion set $\mathcal{C}^\prime$ is constructed by \cite{GOPECCV14} and post-processing. Secondly, we pass $x^\prime$ and $\mathcal{C}^\prime$ through our MIE network to obtain the embedding vectors, $f(x^{\prime c})$. Then, for any text labels $y^\prime$ in the embedding space $\mathcal{S}$, the distance between image $x^\prime$ and $y^\prime$ is computed by $\min_{c\in \mathcal{C^\prime}}D_{f(x^{\prime c}),y^\prime} $. Finally, the corresponding labels of an image is selected according to the distance ranking list. In addition, given a predicted label $y^*$, we can locate the corresponding image subregion $c^*$, \ie, $c^*=\argmin_{c\in\mathcal{C}}D_{f(x_i^c), y^*}$.

Intuitively, in our inference stage, we embed image subregions into the semantic space. And if there is any subregion close enough to a text label, we reckon that this label is associated with the image since there is a corresponding subregion that interprets the label concept, thus such label can be assigned to the image.

\section{Experiments}
To evaluate the proposed model, we conduct experiments on two tasks: multi-label image annotation and zero-shot learning.  

\subsection{Datasets and implementation details}
For the task of multi-label image annotation, we test on the largest publicly available multi-label image dataset, NUS-WIDE \cite{nuswide09}. This dataset contains 209,347 images from Flickr that have been manually annotated. We follow the train-test split of \cite{YunchaoICLR14} to use a subset of 150,000 images for training and the rest for testing. The label dictionary of this dataset contains 81 different labels. 

For the task of zero-shot learning, we train our model on NUS-WIDE dataset and test it on the validation images of Places205 dataset \cite{Places205}. We test on 197 classes which are not included in NUS-WIDE (we exclude the images from 8 classes, \ie, \emph{bridge}, \emph{castle}, \emph{harbor}, \emph{mountain}, \emph{ocean}, \emph{sky}, \emph{tower}, and \emph{valley}, since they are included in NUS-WIDE), and there are 100 validation images per class.

We use Caffe \cite{caffe14} to implement our model. The optimization of our network is achieved by stochastic gradient descent
with a momentum term of weight 0.9 and with mini-batch size of 100. The initial learning rate is set to 0.1, and we update it with the ``steps" policy. A weight decay of 0.0005 is applied.

\subsection{Experiments on multi-label image annotation}
For the task of multi-label image annotation, CNN-based method \cite{YunchaoICLR14} recently reported state-of-the-art performance in NUS-WIDE. Thus we follow the evaluation protocols in \cite{YunchaoICLR14} to evaluate our performance.

\begin{table*}
\centering
\small
\rowcolors{2}{}{gray!35}
\begin{tabular}{ c | c c c c c }
\toprule[0.2 em] %
method& per-class recall & per-class precision & overall recall & overall precision & $N_+$  \\
\midrule
 Upper bound & 97.00&44.87&82.76&66.49&100.00 \\
 \midrule
       CNN + Ranking\cite{YunchaoICLR14} &26.83&31.93 & 58.00& 46.59& 95.06 \\ 
       CNN + WARP\cite{YunchaoICLR14} & 35.60& 31.65& 60.49 & 48.59& 96.29\\ 
       \midrule
       Embed + Ranking &31.59&34.75 &60.26 &49.17 & 98.77\\ 
       MIE + 36 subregions w/o Rank Optimization& 34.71&35.92 &61.87 & 50.53& 98.77\\ 
       MIE w/o Rank Optimization & 38.90&\bf{37.87} & 63.12& 51.55& 98.77\\  
       MIE Full Model & \bf{40.15}& 37.74& \bf{65.03}& \bf{52.23}&\bf{100.00} \\ 
\bottomrule[0.1 em]
\end{tabular}
\caption{Image annotation results on NUS-WIDE with $k$=3 annotated labels per image. See Sec. \ref{Quantitative1} for the definition of ``Upper bound".}
    \label{Exp1Tb1}
\end{table*}

\begin{table*}
\centering
\small
\rowcolors{2}{}{gray!35}
\begin{tabular}{ c | c c c c c }
\toprule[0.2 em] %
method& per-class recall & per-class precision & overall recall & overall precision & $N_+$  \\
\midrule
 Upper bound & 99.57&28.83&96.40&46.22&100.00 \\
 \midrule
       CNN + Ranking\cite{YunchaoICLR14} &42.48&22.74 & 72.78& 35.08& 97.53 \\ 
       CNN + WARP\cite{YunchaoICLR14} & 52.03& 22.31& 75.00 & 36.16& \bf{100.00}\\ 
       \midrule
       Embed + Ranking & 50.25&26.08 & 75.62& 36.94&98.77\\ 
       MIE + 36 subregions w/o Rank Optimization & 53.92&26.83 &76.81 & 37.78&\bf{100.00}\\
       MIE w/o Rank Optimization & 57.79 & 28.19& 79.16& 38.14 &\bf{100.00}\\  
       MIE Full Model & \bf{59.81}& \bf{28.26}& \bf{80.94}&\bf{39.00} & \bf{100.00}\\ 
\bottomrule[0.1 em]
\end{tabular}
\caption{Image annotation results on NUS-WIDE with $k$=5 annotated labels per image. See Sec. \ref{Quantitative1} for the definition of ``Upper bound".}
    \label{Exp1Tb2}
    \vspace{-5pt}
\end{table*}

\begin{figure*}
  \centering
  \includegraphics[width=0.99\textwidth] {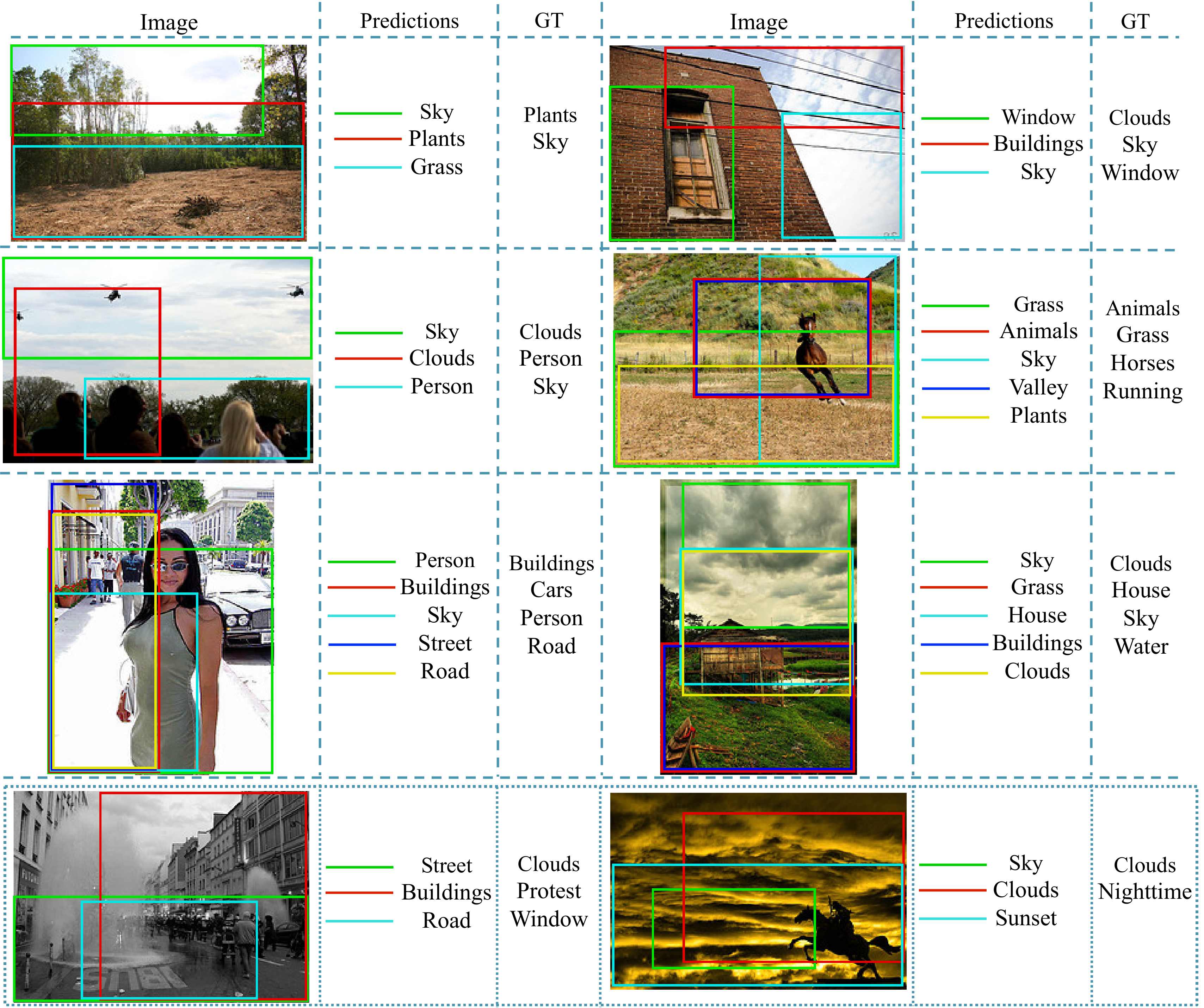}
\caption{ The image annotation results and the discovered image region associated with each label using our method. The first three rows show correct results and the last row shows two failed cases. The predicted labels are listed according to the ranking (We show top-3 predictions if the number of ground truth labels is smaller than or equal to 3, otherwise we show top-5 predictions.). The ground truth labels (GT) are listed according to alphabetic order. Better viewed in color.}
\label{Fig4}
\end{figure*}

\subsubsection{Evaluation Protocols}

\label{Exp1Metric}

For each image, we assign $k$ (e.g., $k$ = 3, 5) highest-ranked labels to the image and compare
the assigned labels to the ground-truth labels. We compute the recall and precision for each label separately, and then report the per-class recall and per-class precision:
\vspace{-10pt}

\begin{small}
\begin{equation}
\text{per-class recall} = \frac{1}{t}\sum_{i=1}^t\frac{N_i^c}{N_i^g},\ \ \ \text{per-class precision} = \frac{1}{t}\sum_{i=1}^t\frac{N_i^c}{N_i^p}, \nonumber
\label{metric1}
\end{equation}
\end{small}
$\!$\noindent where $t$ is the total number of labels, $N_i^c$ is the number of correctly annotated image for label $i$, $N_i^g$ is the number of ground-truth labeling for label $i$, and $N_i^p$ is the number of predictions for label $i$. The above evaluations are biased toward infrequent labels, because making them correct would have a very significant impact on the final accuracy. Therefore we also report the overall recall and overall precision:
\vspace{-15pt}

\begin{small}
\begin{equation}
\text{overall recall} = \frac{\sum_{i=1}^tN_i^c}{\sum_{i=1}^tN_i^g},\ \ \ \text{overall precision} =\frac{\sum_{i=1}^tN_i^c}{\sum_{i=1}^tN_i^p}, \nonumber
\label{metric1}
\end{equation}
\end{small}

Finally, the percentage of recalled labels in all labels is also evaluated, represented as $N_+$. We believe that evaluating all of these 5 metrics makes the evaluation less biased and thorough.

\subsubsection{Quantitative results}
\label{Quantitative1}

To better justify the superiority of our method for modeling the region-to-label correspondence and optimizing the ranking, we test our MIE model with four variants: I) ``Embed + Ranking" which represents our ranking loss embedding baseline defined in Section \ref{LearnEmbeddingF}; II) ``MIE + 36 subregions w/o Rank Optimization" where the subregion set is constructed manually by selecting subregions with minimum side length as 2, in the 4$\times$4 rigidly defined image grid (totally 36 such subregions) and rank optimization in Section \ref{rankoptimize} is not included; III) ``MIE w/o Rank Optimization" as defined in Eq.\ref{MIEloss}, which uses region proposals method \cite{GOPECCV14} to construct subregion set but does not include rank optimization; IV) ``MIE Full Model" that represents our full model in Eq.\ref{MIEloss2} with region proposals and the rank optimization scheme. 

And we also compare our method with two models in \cite{YunchaoICLR14}: 1) ``CNN + Ranking" which represents their ranking loss baseline; 2) ``CNN + WARP" that represents their WARP loss model which is the state-of-the-art model in multi-label image annotation.
We evaluate the performance w.r.t. the number of predicted labels $k$, and the results with $k=3$, $k=5$ are shown in Table \ref{Exp1Tb1}, \ref{Exp1Tb2}, respectively.

In the first row of Table \ref{Exp1Tb1} and \ref{Exp1Tb2}, we list the ``Upper bound" results. Because in NUS-WIDE, each image has different number of ground-truth labels. Thus, we follow \cite{YunchaoICLR14} to define the ``upper bound" method: for each image, when the number of ground-truth labels is larger than $k$, we randomly choose $k$ ground-truth labels and assign them to the image; when the number of ground-truth labels is smaller than $k$, we assign all ground-truth labels to the image and randomly pick other labels to include. This method represents the ``upper bound" performance.

Our ``Embed + Ranking" baseline and the ``CNN + Ranking" baseline in \cite{YunchaoICLR14} are comparable since ranking loss is used in both methods. As we see, in the second and fourth rows of the Tables, ``Embed + Ranking" outperforms ``CNN + Ranking" by 3.23\% averaged in all metrics for $k=3$ and 3.41\% for $k=5$. One reason for this is we use Googlenet to obtain image features and \cite{YunchaoICLR14} used Alexnet.

Thus, for fair comparison, we compare with our ranking loss baseline ``Embed + Ranking" model. Overall, our ``MIE Full Model" outperforms the ``Embed + Ranking" baseline by 4.12\% averaged in all metrics for $k=3$ and 4.07\% for $k=5$. Several factors may contribute to such improvement.

We compare various variants of our method and the ``Embed + Ranking" model to justify the contribution of several key components we proposed. Firstly, ``MIE w/o Rank Optimization" introduces the idea of modeling region-to-label correspondence. As shown in the sixth row of the Tables, it outperforms ``Embed + Ranking"  baseline by 3.13\% for $k=3$ and 3.12\% for $k=5$, which validates the contribution of modeling region-to-label correspondence. Then, by considering rank optimization, our ``MIE Full Model" further boost the performance of ``MIE w/o Rank Optimization" by 0.99\% for $k=3$ and 0.95\% for $k=5$. Besides, the effectiveness of constructing image subregion set using region proposals method is evaluated. In the fifth and sixth rows of the Tables, we compare the method with region proposals to the one that manually generates subregions, which shows an improvement of 1.68\% for $k=3$ and 1.59\% for $k=5$.

And finally, our ``MIE Full Model" outperforms the state-of-the-art method ``CNN + WARP" in all metrics by a large margin, of 4.51\% averaged over 5 metrics for $k=3$ and 4.50\% for $k=5$. 

\subsubsection{Qualitative results}
As discussed in Section \ref{inference}, our model is able to discover the image regions corresponding to the predicted labels, \ie, for a predicted label $y^*$, $c^*=\argmin_{c\in\mathcal{C}}D_{f(x_i^c), y^*}$ is its corresponding region. 

In Fig.\ref{Fig4}, we show the image annotation result, as well as the localized corresponding region for each predicted label. The first three rows show some good annotation cases that most of the ground truth labels are correctly predicted. As we see, for these images, the correctly predicted labels are associated with reasonable regions, such as \emph{Sky} of the upper left image, \emph{Window} of the upper right image, \emph{Person} of the middle left image, \emph{Animals} of the middle right image, \emph{Road} of the lower left image, and \emph{Grass} of the lower right image, \etc., are reasonably discovered by the associated bounding boxes. There are some wrong annotations or inaccurate bounding boxes as well. For instance, the middle right image are annotated with \emph{Plants}. But if we look at the bounding box of \emph{Plants}, the subregion interprets the label concepts well. The same case happens in the last row of Fig.\ref{Fig4} where two failed cases are shown. Even the predicted \emph{Buildings} of the left image and \emph{Sunset} of the right image are not included in the ground truth labels, the subregions in the bounding boxes interpret the predicted labels very well. It is important to note that our task is not detection and no bounding box annotation is provided in training. Thus, some objects are not tightly bounded by the boxes, \eg, \emph{Animals} of the right image in the second row. Overall, our method is able to correctly annotate images with labels, and more importantly, the semantic region associated with each label can be located, which is a distinct advantage of our method over existing multi-label annotation methods.

\begin{figure*}
  \centering
  \includegraphics[width=0.99\textwidth] {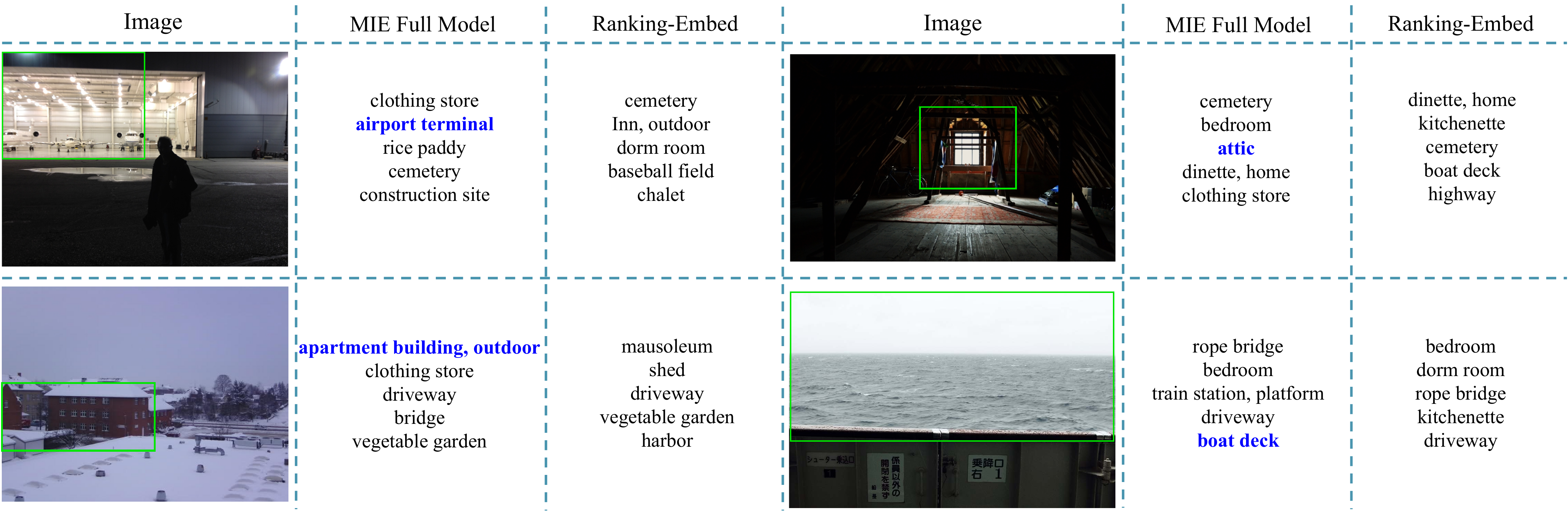}
\caption{ Zero-shot test images from Places205, and their corresponding top-5 labels predicted by MIE Full Model and Embed + Ranking baseline model. All the labels predicted are not seen during training stage for both models. The correct labels are shown in blue. The semantically-meaningful image subregion associated with each correctly predicted label are shown in green bounding boxes.}
\label{Fig5}
\end{figure*}

\subsection{Experiments on zero-shot learning}
Another distinct advantage of our model is its ability to make reasonable inferences about candidate labels
that it has never visually seen at training stage. For example, a MIE model trained on images labeled \emph{swallow}, \emph{magpie}, and \emph{woodpecker} would likely have the ability to generalize to other labels related to bird, because the language model encodes the general concept of bird which share similar visual appearance. Thus if tested on images of highly specific classes of bird which the model has never seen before, for example a photo of an \emph{pelican}, and asked whether the correct label is more likely \emph{pelican} or some other unfamiliar label, say $panda$, our model stands a fighting chance of guessing correctly because the language model ensures that the embedding of $pelican$ is closer to that of birds our model has seen, while the embedding of \emph{panda} is closer to those of mammals.

\begin{table}
\centering
\scriptsize
\rowcolors{2}{}{gray!35}
\begin{tabular}{ c | c c c c }
\toprule[0.2 em] %
method& MAP@1 & MAP@2 & MAP@5 & MAP@10 \\
\midrule
Embed + Ranking &  6.53 &  10.18 &  18.92 & 28.17  \\
       MIE Full Model & \bf{7.14} & \bf{11.29} & \bf{20.50} & \bf{30.27}\\ 
\bottomrule[0.1 em]
\end{tabular}
\caption{Zero-shot learning results on 197 classes of Places205 dataset.}
    \label{exp2table3}
\end{table}
To evaluate the scalability of our model w.r.t. the number of label categories, we train our model on NUS-WIDE dataset and use it to predict the label of images from the 197 previously unseen labels in Places205 dataset.

Fig.\ref{Fig5} depicts some qualitative results. The two columns by the side of each image show the top-5 label predictions of MIE model and the ranking loss baseline model defined in Section \ref{LearnEmbeddingF}. The superiority of visual-semantic models over classification methods in zero-shot learning has been thoroughly validated in \cite{DeviSE}. Since classification methods need to retrain the classifiers whenever a new class emerges, they could not generate to new labels. Thus we do not test with classification methods.

``MIE Full Model" and ``Embed + Ranking" are visual-semantic models that utilize information from the label semantic space, thus, as it shows, both methods predict reasonable results. Even ``Embed + Ranking" does not predict the true labels for those images,  its predictions are generally semantically ``close" to the desired labels.

 MIE successfully predicts a wide range of labels outside its training set. In addition, the semantically-meaningful image subregion associated with each correctly predicted label are shown in Fig.\ref{Fig5}. As we see, the localized image region interprets the label concept in some aspects. For instance, for the upper right image, \emph{attic} is semantically close to \emph{window} from the NUS-WIDE dataset, thus a window region is discovered and the semantic concept of such specific region is transferred to assist the prediction of new labels.
 
To quantify the performance, the mean average precision@$k$ (MAP@$k$) metric is adopted, which measures the precision of predicting the ground truth label within the top-$k$ predicted labels. The results are shown in Table \ref{exp2table3}. As we see, our MIE model outperforms ranking loss baseline model in all metrics by 1.35\% on average.

Taken together, the multi-label image annotation and zero-shot learning experiments validate the effectiveness of modeling the region-to-label correspondence and optimizing the ranking for visual-semantic embedding models. Moreover, the proposed model could be applied to various visual tasks, especially the ones related to images and texts. 

\section{Conclusion and Future Work}
In this paper, we have proposed a novel multi-instance visual-semantic embedding model for general images with single or multiple labels. Instead of embedding a whole image into the semantic space, we propose to embed the semantically-meaningful image subregions to the corresponding image labels. Such an embedding space possesses superior representation and generalization power over previous methods. By experimental validation, we show that: 1) Our model not only outperforms state-of-the-art method in multi-label image annotation, but also can discover the corresponding region that interprets the concept of each image label; 2) Our model has the ability to make correct predictions across plenty of previously unseen labels by leveraging semantic knowledge elicited from the embedding space.
In the future, we would like to explore the modeling in the semantic space, and apply our model to various vision applications.

{\small
\bibliographystyle{ieee}
\bibliography{egbib}

\begin{thebibliography}{10}\itemsep=-1pt

\bibitem{ChenBOWCVPR12}
Q.~Chen, Z.~Song, Y.~Hua, Z.~Huang, and S.~Yan.
\newblock Hierarchical matching with side information for image classification.
\newblock In {\em CVPR}, 2012.

\bibitem{nuswide09}
T.-S. Chua, J.~Tang, R.~Hong, H.~Li, Z.~Luo, and Y.-T. Zheng.
\newblock Nus-wide: a real-world web image database from national university of
  singapore.
\newblock In {\em ACM International Conference on Image and Video Retrieval},
  2009.

\bibitem{JiaDengECCV14}
J.~Deng, N.~Ding, Y.~Jia, A.~Frome, K.~Murphy, S.~Bengio, Y.~Li, H.~Neven, and
  H.~Adam.
\newblock Large-scale object classification using label relation graphs.
\newblock In {\em ECCV}, 2014.

\bibitem{Imagenet09}
J.~Deng, W.~Dong, R.~Socher, L.~Li, K.~Li, and L.~Fei-Fei.
\newblock Imagenet: a large-scale hierachical image database.
\newblock In {\em CVPR}, 2009.

\bibitem{attribute1}
A.~Farhadi, I.~Endres, D.~Hoiem, and D.~Forsyth.
\newblock Describing objects by their attributes.
\newblock In {\em CVPR}, 2009.

\bibitem{DeviSE}
A.~Frome, G.~Corrado, J.~Shlens, S.~Bengio, J.~Dean, M.~Ranzato, and
  T.~Mikolov.
\newblock Devise: A deep visual-semantic embedding model.
\newblock In {\em NIPS}, 2013.

\bibitem{fastrcnn15}
R.~Girshick.
\newblock Fast r-cnn.
\newblock In {\em ICCV}, 2015.

\bibitem{RCNN14}
R.~Girshick, J.~Donahue, T.~Darrell, and J.~Malik.
\newblock Rich feature hierarchies for accurate object detection and semantic
  segmentation.
\newblock In {\em CVPR}, 2014.

\bibitem{YunchaoICLR14}
Y.~Gong, Y.~Jia, T.~K. Leung, A.~Toshev, and S.~Ioffe.
\newblock Deep convolutional ranking for multilabel image annotation.
\newblock In {\em ICLR}, 2014.

\bibitem{Tagprop09}
M.~Guillaumin, T.~Mensink, J.~Verbeek, and C.~Schmid.
\newblock Tagprop: Discriminative metric learning in nearest neighbor models
  for image auto-annotation.
\newblock In {\em ICCV}, 2009.

\bibitem{caffe14}
Y.~Jia, E.~Shelhamer, J.~Donahue, S.~Karayev, J.~Long, R.~Girshick,
  S.~Guadarrama, and T.~Darrell.
\newblock Caffe: Convolutional architecture for fast feature embedding.
\newblock {\em arXiv preprint arXiv:1408.5093}, 2014.

\bibitem{GOPECCV14}
P.~Kr{\"a}henb{\"u}hl and V.~Koltun.
\newblock Geodesic object proposals.
\newblock In {\em ECCV}, 2014.

\bibitem{Alexnet12}
A.~Krizhevsky, I.~Sutskever, and G.~Hinton.
\newblock Imagenet classification with deep convolutional neural networks.
\newblock In {\em NIPS}, 2012.

\bibitem{LecunNIPS90}
Y.~LeCun, B.~Boser, J.~Denker, D.~Henderson, R.~Howard, W.~Hubbard, and
  L.~Jackel.
\newblock Handwritten digit recognition with a back-propagation network.
\newblock In {\em NIPS}, 1990.

\bibitem{MakadiaECCV08}
A.~Makadia, V.~Pavlovic, and S.~Kumar.
\newblock A new baseline for image annotation.
\newblock In {\em ECCV}, 2008.

\bibitem{Word2VecNIPS13}
T.~Mikolov, I.~Sutskever, K.~Chen, G.~Corrado, and J.~Dean.
\newblock Distributed representations of words and phrases and their
  compositionality.
\newblock In {\em NIPS}, 2013.

\bibitem{NoriuziICLR14}
M.~Norouzi, T.~Mikolov, S.~Bengio, Y.~Singer, J.~Shlens, A.~Frome, G.~Corrado,
  and J.~Dean.
\newblock Zero-shot learning by convex combination of semantic embeddings.
\newblock In {\em ICLR}, 2014.

\bibitem{marknips09}
M.~Palatucci, D.~Pomerleau, G.~E. Hinton, and T.~M. Mitchell.
\newblock Zero-shot learning with semantic output codes.
\newblock In {\em NIPS}, 2009.

\bibitem{relativeattr11}
D.~Parikh and K.~Grauman.
\newblock Relative attributes.
\newblock In {\em ICCV}, 2011.

\bibitem{glove14}
J.~Pennington, R.~Socher, and C.~D. Manning.
\newblock Glove: Global vectors for word representation.
\newblock In {\em EMNLP}, 2014.

\bibitem{FisherVectorCVPR07}
F.~Perronnin and C.~Dance.
\newblock Fisher kernels on visual vocabularies for image categorization.
\newblock In {\em CVPR}, 2007.

\bibitem{ZRICCV15}
Z.~Ren, C.~Wang, and A.~Yuille.
\newblock Scene-domain active part models for object representation.
\newblock In {\em ICCV}, 2015.

\bibitem{marcuscvpr11}
M.~Rohrbach, M.~Stark, and B.~Schiele.
\newblock Evaluating knowledge transfer and zero-shot learning in a large-scale
  setting.
\newblock In {\em CVPR}, 2011.

\bibitem{vggnet15}
K.~Simonyan and A.~Zisserman.
\newblock Very deep convolutional networks for large-scale image recognition.
\newblock In {\em ICLR}, 2015.

\bibitem{sochernips13}
R.~Socher, M.~Ganjoo, H.~Sridhar, O.~Bastani, C.~D. Manning, and A.~Y. Ng.
\newblock Zero-shot learning through cross-modal transfer.
\newblock In {\em NIPS}, 2013.

\bibitem{googlenet15}
C.~Szegedy, W.~Liu, Y.~Jia, P.~Sermanet, S.~Reed, D.~Anguelov, D.~Erhan,
  V.~Vanhoucke, and A.~Rabinovich.
\newblock Going deeper with convolutions.
\newblock In {\em CVPR}, 2015.

\bibitem{wsabie11}
J.~Weston, S.~Benjio, and N.~Usunier.
\newblock Wsabie: scaling up to large vocabulary image annotation.
\newblock In {\em IJCAI}, 2011.

\bibitem{JianchaoCVPR09}
J.~Yang, K.~Yu, Y.~Gong, and T.~Huang.
\newblock Linear spatial pyramid matching using sparse coding for image
  classification.
\newblock In {\em CVPR}, 2009.

\bibitem{visualCNN14}
M.~Zeiler and R.~Fergus.
\newblock Visualizing and understanding convolutional networks.
\newblock In {\em ECCV}, 2014.

\bibitem{Places205}
B.~Zhou, A.~Lapedriza, J.~Xiao, A.~Torralba, and A.~Oliva.
\newblock Learning deep features for scene recognition using places database.
\newblock In {\em NIPS}, 2014.

\end{thebibliography}
}

\end{document}